\pdfoutput=1
\documentclass[10pt,twocolumn,letterpaper]{article}

\usepackage[pagenumbers]{cvpr} %

\usepackage{graphicx}
\usepackage{amsmath}
\usepackage{amssymb}
\usepackage[table]{xcolor}
\usepackage[shortlabels]{enumitem}
\usepackage{booktabs}
\usepackage{multirow}

\DeclareFontFamily{U}{mathx}{\hyphenchar\font45}
\DeclareFontShape{U}{mathx}{m}{n}{
      <5> <6> <7> <8> <9> <10>
      <10.95> <12> <14.4> <17.28> <20.74> <24.88>
      mathx10
      }{}
\DeclareSymbolFont{mathx}{U}{mathx}{m}{n}
\DeclareFontSubstitution{U}{mathx}{m}{n}
\DeclareMathAccent{\widecheck}{0}{mathx}{"71}

\usepackage[pagebackref,breaklinks,colorlinks]{hyperref}

\usepackage[capitalize]{cleveref}
\crefname{section}{Sec.}{Secs.}
\Crefname{section}{Section}{Sections}
\Crefname{table}{Table}{Tables}
\crefname{table}{Tab.}{Tabs.}

\newcommand{\bs}[1]{\boldsymbol{#1}}
\newcommand{\mb}[1]{\boldsymbol{#1}}
\newcommand{\wh}[1]{\widehat{#1}}
\newcommand{\wc}[1]{\widecheck{#1}}
\newcommand{\wt}[1]{\widetilde{#1}}
\newcommand{\rot}{\gamma}

\definecolor{grassgreen}{RGB}{122,201,67}

\def\confName{CVPR}
\def\confYear{2022}

\begin{document}

\title{GLAMR: Global Occlusion-Aware Human Mesh Recovery \\ with Dynamic Cameras}

\author{
Ye Yuan\textsuperscript{2\thanks{Work done during an internship at NVIDIA.}} \qquad Umar Iqbal\textsuperscript{1} \qquad Pavlo Molchanov\textsuperscript{1} \qquad Kris Kitani\textsuperscript{2} \qquad Jan Kautz\textsuperscript{1} \\[1mm]
\textsuperscript{1}NVIDIA \qquad \textsuperscript{2}Carnegie Mellon University \\[1mm]
{ \url{https://nvlabs.github.io/GLAMR}} \\
}
\maketitle

\begin{abstract}
\vspace{-2mm}
We present an approach for 3D global human mesh recovery from monocular videos recorded with dynamic cameras. Our approach is robust to severe and long-term occlusions and tracks human bodies even when they go outside the camera's field of view. To achieve this, we first propose a deep generative motion infiller, which autoregressively infills the body motions of occluded humans based on visible motions. Additionally, in contrast to prior work, our approach reconstructs human meshes in consistent global coordinates even with dynamic cameras. Since the joint reconstruction of human motions and camera poses is underconstrained, we propose a global trajectory predictor that generates global human trajectories based on local body movements. Using the predicted trajectories as anchors, we present a global optimization framework that refines the predicted trajectories and optimizes the camera poses to match the video evidence such as 2D keypoints. Experiments on challenging indoor and in-the-wild datasets with dynamic cameras demonstrate that the proposed approach outperforms prior methods significantly in terms of motion infilling and global mesh recovery.
\end{abstract}

\vspace{-4mm}
\section{Introduction}
\label{sec:intro}
\vspace{-1mm}

Recovering fine-grained 3D human meshes from monocular videos is essential for understanding human behaviors and interactions, which can be the cornerstone for numerous applications including virtual or augmented reality, assistive living, autonomous driving, \etc. Many of these applications use dynamic cameras to capture human behaviors yet also require estimating human motions in global coordinates consistent with their surroundings. For instance, assistive robots and autonomous vehicles need a holistic understanding of human behaviors and interactions in the world to safely plan their actions even when they are moving. Therefore, our goal in this paper is to tackle the important task of recovering global human meshes from monocular videos captured by dynamic cameras.

However, this task is highly challenging for two main reasons. First, dynamic cameras make it difficult to estimate human motions in \emph{consistent global coordinates}. Existing human mesh recovery methods estimate human meshes in the camera coordinates~\cite{moon2019camera,zhen2020smap} or even in the root-relative coordinates~\cite{kocabas2020vibe,moon2020i2l}. Hence, they can only recover global human meshes from dynamic cameras by using SLAM to estimate camera poses~\cite{liu20204d}. However, SLAM can often fail for in-the-wild videos due to moving and dynamic objects. It also has the problem of scale ambiguity, which often leads to camera poses that are inconsistent with the human motions. Second, videos captured by dynamic cameras often contain \emph{severe and long-term occlusions} of humans, which can be caused by missed detection, complete obstruction by objects and other people, or the person going outside the camera's field of view (FoV). These occlusions pose serious challenges to standard human mesh recovery methods, which rely on detections or visible parts to estimate human meshes. Only a few works have attempted to tackle the occlusion problem in human mesh recovery~\cite{jiang2020coherent,fieraru2020three}. However, these methods can only address partial occlusions of a person and fail to handle severe occlusions when the person is completely invisible for an extended period of time.

To tackle the above challenges, we propose Global Occlusion-Aware Human Mesh Recovery (GLAMR), which can handle severe occlusions and estimate human meshes in consistent global coordinates -- even for videos recorded with dynamic cameras. We start by using off-the-shelf methods (\eg, KAMA~\cite{iqbal2021kama} or SPEC~\cite{Kocabas_SPEC_2021}) to estimate the shape and pose sequences (motions) of visible people in the camera coordinates. These methods also rely on multi-object tracking and re-identification, which provide occlusion information, and the motion of occluded frames is not estimated. To tackle potentially severe occlusions, we propose a deep generative motion infiller that autoregressively infills the local body motions of occluded people based on visible motions. The motion infiller leverages human dynamics learned from a large motion database, AMASS~\cite{AMASS:ICCV:2019}. Next, to obtain global motions, we propose a global trajectory predictor that can generate global human trajectories based on local body motions. It is motivated by the observation that the global root trajectory of a person is highly correlated with the local body movements. Finally, using the predicted trajectories as anchors to constrain the solution space, we further propose a global optimization framework that jointly optimizes the global motions and camera poses to match the video evidence such as 2D keypoints.

The contributions of this paper are as follows: \textbf{(1)} We propose the first approach to address long-term occlusions and estimate global 3D human pose and shape from videos captured by dynamic cameras; \textbf{(2)} We propose a novel generative Transformer-based motion infiller that autoregressively infills long-term missing motions, which considerably outperforms state-of-the-art motion infilling methods; \textbf{(3)} We propose a method to generate global human trajectories from local body motions and use the generated trajectories as anchors to constrain global motion and camera optimization; \textbf{(4)} Extensive experiments on challenging indoor and in-the-wild datasets demonstrate that our approach outperforms prior state-of-the-art methods significantly in tackling occlusions and estimating global human meshes.

\begin{figure*}[t]
\centering
\includegraphics[width=\linewidth]{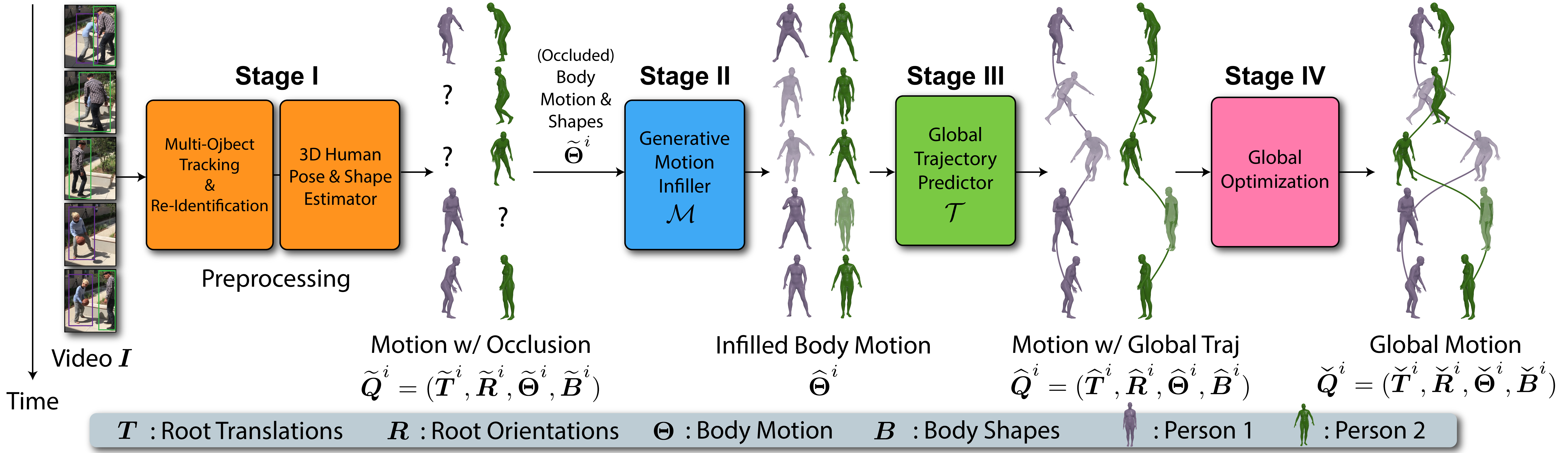}
\vspace{-6mm}
\caption{\textbf{Overview} of GLAMR. In \textbf{Stage I}, we preprocess the video with multi-object tracking, re-identification and human mesh recovery to extract each person's occluded motion $\mb{\wt{Q}}^i$ in the camera coordinates. In \textbf{Stage II}, we propose a generative motion infiller to infill the occluded body motion $\mb{\wt{\Theta}}^i$ to produce occlusion-free body motion $\mb{\wh{\Theta}}^i$. In \textbf{Stage III}, we propose a global trajectory predictor that uses the infilled body motion $\mb{\wh{\Theta}}^i$ to generate the global trajectory $(\mb{\wh{T}}^i,\mb{\wh{R}}^i)$ of each person and obtain their global motion $\mb{\wh{Q}}^i$. In \textbf{Stage IV}, we jointly optimize the global trajectories of all people and the camera parameters to produce global motions $\mb{\wc{Q}}^i$ consistent with the video.}
\label{fig:overview}
\vspace{-1mm}
\end{figure*}

\section{Related Work}
\label{sec:related_work}
\noindent\textbf{Camera-Relative Pose Estimation.}
3D human mesh recovery from RGB images or videos is an ill-posed problem due to the depth ambiguity. Most  existing methods simplify the problem by estimating human poses relative to the pelvis (root) of the human body~\cite{Akhter:CVPR:2015, bogo2016keep, lassner2017unite, hmrKanazawa18, pavlakos2018humanshape, guler2019holo, kolotouros2019spin, pavlakos2019texture, Rong_2019_ICCV, kolotouros2019convolutional, choutas2020expose, zanfir2020weakly, sun2019human, joo2021eft, choi2020pose, kundu2020mesh, SMPL-X:2019, xu2019denserac, monototalcapture2019, song2020human, zhang2020object, zhou2021monocular, moon2020i2l, lin2021end, Mueller:CVPR:21, kolotouros2021prohmr, Zhang_2021_ICCV, Sun_2021_ICCV, humanMotionKanazawa19, kocabas2020vibe, luo20203d, choi2020beyond, rempe2021humor}. These methods assume an orthographic camera projection model and neglect the absolute 3D translation of the person \wrt the camera. To address the lack of translation, recent methods start to estimate human meshes in the camera coordinates~\cite{zanfir2018monocular, jiang2020coherent, Zanfir_2021_ICCV, ICG, Zhang_2021_CVPR, Xie_2021_ICCV, PhysCapTOG2020, liu20204d, li2020hybrik, iqbal2021kama, reddy2021tesstrack}. Several approaches recover the absolute translation of the person using an optimization framework~\cite{mono20173dhp, mehta2017vnect, XNect_SIGGRAPH2020,zanfir2018deep, rogez2017lcr}. A few methods exploit various scene constraints during the optimization process to improve depth prediction~\cite{zanfir2018monocular,Weng_2021_CVPR}. Alternatively, recent approaches use physics-based constraints to ensure the physical plausibility of the estimated poses~\cite{PhysCapTOG2020, Xie_2021_ICCV, GraviCap2021, yuan2021simpoe, isogawa2020optical}. Iqbal~\etal~\cite{iqbal2020learning} exploit a limb-length constraint to recover the absolute translation of the person using a 2.5D representation. Some approaches approximate the depth of the person using the bounding box size~\cite{jiang2020coherent, moon2019camera, Zhang_2021_CVPR}. HybrIK~\cite{li2020hybrik} and KAMA~\cite{iqbal2021kama} employ inverse kinematics to estimate human meshes with absolute translations in the camera coordinates. Several methods directly predict the absolute depth of each person using a heatmap representation~\cite{Fabbri_2020_CVPR,zhen2020smap}. Recently, SPEC~\cite{Kocabas_SPEC_2021} learns to predict the camera parameters (pitch, yaw, FoV) from the image, which are used for absolute pose regression in the camera coordinates. THUNDR~\cite{Zanfir_2021_ICCV} also adopts a similar strategy but uses known camera parameters. While these methods show impressive results, they cannot estimate global human motions from videos captured by dynamic cameras. In contrast, our approach can recover human meshes in consistent global coordinates for dynamic cameras and handle severe and long-term occlusions.

\vspace{2mm}
\noindent\textbf{Global Pose Estimation.}
Most existing methods that estimate 3D poses in world coordinates rely on calibrated, synchronized, and static multi-view capture setups~\cite{belagiannis20143d,joo2018total,reddy2021tesstrack, multiviewpose, zhang20204d, dong2021shape, zhang2021lightweight, zheng2021deepmulticap, huang2021dynamic, dong2021fastpami}. Huang~\etal~\cite{wang2021dynamic} use uncalibrated cameras but still assume time synchronization and static camera setups. Hasler~\etal~\cite{hasler2009markerless} handle unsynchronized moving cameras but assume multi-view input and rely on audio stream for synchronization.  More recently, Dong~\etal~\cite{dong2020motion} propose to recover 3D poses from unaligned internet videos of different actors performing the same activity from unknown cameras. However, they assume that multiple viewpoints of the same pose are available in the videos. Different from these methods, our approach estimates human meshes in global coordinates from \emph{monocular} videos recorded with dynamic cameras. Several methods rely on additional IMU sensors or pre-scanned environments to recover global human motions~\cite{vonMarcard2018,hps2021Vladmir}, which is unpractical for large-scale adoption. Recently, another line of work starts to focus on estimating accurate human-scene interaction~\cite{hassan2019resolving,luo2021dynamics,yi2022human,huang2022cap}. Liu~\etal~\cite{liu20204d} first obtain the camera poses and dense reconstruction of the scene from dynamic cameras using a SLAM algorithm, COLMAP~\cite{schonberger2016structure}. The camera poses are used for camera-to-world transformation, while the reconstructed scene is used to encourage human-scene contacts. However, SLAM can often fail for the in-the-wild videos and is prone to error propagation. In contrast, our approach does not require SLAM but instead uses global trajectory prediction to constrain the joint reconstruction of human motions and camera poses. Additionally, our approach can also handle severe and long-term occlusions common in dynamic camera setups.

\vspace{2mm}
\noindent\textbf{Occlusion-Aware Pose Estimation.}
Most existing human pose estimation methods assume the person is fully visible in the images and are not robust to strong occlusions. Only a few methods address the occlusion problem in pose estimation~\cite{zhang2020object,rockwell2020fullbody,fieraru2020three,rempe2021humor,pare2021kocabas}. While these methods show impressive results under partial occlusions, they do not address severe and long-term occlusions when people are completely obstructed or outside the camera's FoV for a long time. In contrast, our approach leverages deep generative human motion models to tackle severe and long-term occlusions.

\vspace{2mm}
\noindent\textbf{Human Motion Modeling.}
Extensive research has studied 3D human dynamics for various tasks including motion prediction and synthesis~\cite{fragkiadaki2015recurrent,jain2016structural,li2017auto,martinez2017human,villegas2017learning,pavllo2018quaternet,aksan2019structured,gopalakrishnan2019neural,yan2018mt,barsoum2018hp,yuan2019diverse,yuan2020dlow,yuan2020residual,cao2020long,petrovich2021action,hassan2021stochastic}. Recent human pose estimation methods start to leverage learned human dynamics models to improve the accuracy of estimated motions~\cite{kocabas2020vibe,rempe2021humor,zhang2021learning}. Several motion infilling approaches are also proposed to generate complete motions from partially observed motions~\cite{hernandez2019human,kaufmann2020convolutional,harvey2020robust,khurana2021detecting}. Additionally, recent work on motion capture shows that global human translations can be predicted from 3D local joint positions~\cite{schreiner2021global}. In contrast to prior work, our trajectory predictor does not require GT root orientations but can predict both global root translations and orientations. Furthermore, we also propose a novel generative autoregressive motion infiller that can use noisy poses as input instead of high-quality GT poses, and we demonstrate its effectiveness in tackling long-term occlusions in human pose estimation.

\section{Method}
\label{sec:method}

\begin{figure*}[t]
\centering
\includegraphics[width=\linewidth]{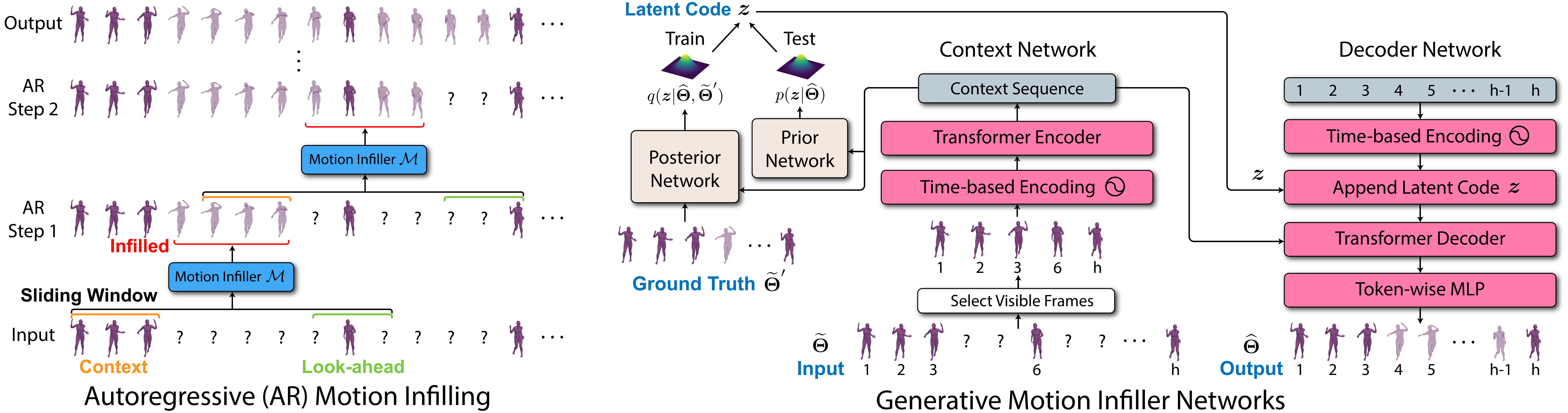}
\vspace{-6mm}
\caption{\textbf{Left:} We autoregressively infill the motion using a sliding window, where the first $h_\texttt{c}$ frames are already infilled to serve as \textcolor{orange}{context} and the last $h_\texttt{l}$ frames are \textcolor{grassgreen}{look-ahead} to guide the ending motion. Frames between the context and look-ahead are \textcolor{red}{infilled}. \textbf{Right:} The CVAE-based motion infiller adopts a Transformer-based seq2seq architecture, where we encode only the visible frames of occluded body motion $\mb{\wt{\Theta}}$ into a context sequence, which is used jointly with latent code $\mb{z}$ by a decoder network to generate occlusion-free motion~$\mb{\wh{\Theta}}$.}
\label{fig:mfiller}
\vspace{-1mm}
\end{figure*}

The input to our framework is a video $\mb{I} = (\mb{I}_1, \ldots, \mb{I}_T)$ with $T$ frames, which is captured by a \emph{dynamic camera}, \ie, the camera poses can change every frame. Our goal is to estimate the global motion (pose sequence) $\{\mb{Q}^i\}_{i=1}^N$ of the $N$ people in the video in a \emph{consistent global coordinate} system. The global motion $\mb{Q}^i= (\mb{T}^i, \bs{R}^i, \bs{\Theta}^i, \mb{B}^i)$ for person $i$ consists of the root translations $\mb{T}^i = (\bs{\tau}^{i}_{s_i}, \ldots, \bs{\tau}^{i}_{e_i})$, root rotations $\mb{R}^i = (\mb{\rot}^{i}_{s_i}, \ldots, \mb{\rot}^{i}_{e_i})$, as well as the body motion $\bs{\Theta}^i = (\bs{\theta}^{i}_{s_i}, \ldots, \bs{\theta}^{i}_{e_i})$ and shapes $\mb{B}^i = (\bs{\beta}^{i}_{s_i}, \ldots, \bs{\beta}^{i}_{e_i})$, where the motion spans from the the first frame $s_i$ to the last frame $e_i$, when the person $i$ is relevant in the video.
In particular, each body pose $\bs{\theta}^{i}_t \in \mathbb{R}^{23 \times 3}$ and shape $\bs{\beta}^{i}_t \in \mathbb{R}^{10}$ corresponds to the pose parameters (excluding root rotation) and shape parameters of the SMPL model~\cite{SMPL:2015}. Using the root translation $\mb{\tau} \in \mathbb{R}^{3}$ and (axis-angle) rotation $\mb{\rot}\in \mathbb{R}^{3}$, SMPL represents a human body mesh with a linear function $\mathcal{S}(\mb{\tau}, \mb{\rot}, \mb{\theta}, \mb{\beta})$ that maps a global pose $\mb{q} = (\mb{\tau}, \mb{\rot}, \bs{\theta}, \bs{\beta})$ to an articulated triangle mesh $\mb{\Phi} \in \mathbb{R}^{K\times 3}$ with $K=6980$ vertices. We can therefore recover the global mesh sequence for each person from their global motion $\mb{Q}^i$ via SMPL.

As outlined in Fig.~\ref{fig:overview}, our framework consists of four stages. In \textbf{Stage I}, we first use multi-object tracking (MOT) and re-identification algorithms to obtain the bounding box sequence of each person, which is input to a human mesh recovery method (\eg, KAMA~\cite{iqbal2021kama} or SPEC~\cite{Kocabas_SPEC_2021}) to extract the motion $\mb{\wt{Q}}^i$ of each person (including translation) in the camera coordinates. The motion $\mb{\wt{Q}}^i$ may be incomplete due to various occlusions (\eg, obstruction, missed detection, going outside FoV), where bounding boxes from MOT are missing for some frames. In \textbf{Stage II} (Sec.~\ref{sec:motion_infill}), we propose a generative motion infiller to tackle the occlusions in the estimated body motion $\mb{\wt{\Theta}}^i$ and produce occlusion-free body motion $\mb{\wh{\Theta}}^i$. In \textbf{Stage III} (Sec.~\ref{sec:traj_pred}), we propose a global trajectory predictor that uses the infilled body motion $\mb{\wh{\Theta}}^i$ to generate the global trajectory (root translations and rotations) of each person and obtain their global motion $\mb{\wh{Q}}^i$. In \textbf{Stage IV} (Sec.~\ref{sec:global_opt}), we jointly optimize the global trajectories of all people and the camera parameters to produce global motions $\mb{\wc{Q}}^i$ consistent with the video evidence.

\subsection{Generative Motion Infiller}
\label{sec:motion_infill}
The task of the generative motion infiller $\mathcal{M}$ is to infill the occluded body motion $\mb{\wt{\Theta}}^i$ of each person to produce occlusion-free body motion $\mb{\wh{\Theta}}^i$. Here, we do not use the motion infiller $\mathcal{M}$ to infill other components in the estimated motion $\mb{\wh{Q}}^i$, \ie, root trajectory ($\mb{\wt{T}}^i,\mb{\wt{R}}^i$) and shapes $\mb{\wt{B}}^i$. This is because it is difficult to infill the root trajectory $(\mb{\wt{T}}^i, \mb{\wt{R}}^i)$ using learned human dynamics, since it resides in the camera coordinates rather than a consistent coordinate system due to the dynamic camera. In Sec.~\ref{sec:traj_pred}, we will use the proposed global trajectory predictor to generate occlusion-free global trajectory $(\mb{\wh{T}}^i,\mb{\wh{R}}^i)$ from the infilled body motion $\mb{\wh{\Theta}}^i$. The trajectory $(\mb{\wt{T}}^i, \mb{\wt{R}}^i)$ from the pose estimator is not discarded and will be used in the global optimization (Sec.~\ref{sec:global_opt}). We use linear interpolation to produce occlusion-free shapes $\mb{\wh{B}}^i$, which can be time-varying to be compatible with per-frame pose estimators such as KAMA.

Given a general occluded human body motion $\bs{\wt{\Theta}} = (\bs{\wt{\theta}}_1, \ldots, \bs{\wt{\theta}}_h)$ of $h$ frames and its visibility mask $\mb{V} = (V_1, \ldots, V_{h})$ as input, the motion infiller $\mathcal{M}$ outputs a complete occlusion-free motion $\bs{\wh{\Theta}} = (\bs{\wh{\theta}}_1, \ldots, \bs{\wh{\theta}}_h)$. The visibility mask $\mb{V}$ encodes the visibility of the occluded motion $\bs{\wt{\Theta}}$, where $V_{t} = 1$ if the body pose $\bs{\wt{\theta}}_t$ is visible in frame $t$ and $V_{t} = 0$ otherwise. Since the human pose for occluded frames can be highly uncertain and stochastic, we formulate the motion infiller $\mathcal{M}$ using the conditional variational autoencoder (CVAE)~\cite{kingma2013auto}:
\begin{align}
\label{eq:motion_infill}
\bs{\wh{\Theta}} = \mathcal{M} (\bs{\wt{\Theta}}, \mb{V}, \mb{z})\,,
\end{align}
where the motion infiller $\mathcal{M}$ corresponds to the CVAE decoder and $\mb{z}$ is a Gaussian latent code. We can obtain different occlusion-free motions $\mb{\wh{\Theta}}$ by varying $\mb{z}$.

\vspace{2mm}
\noindent\textbf{Autoregressive Motion Infilling.}
To ensure that the motion infiller $\mathcal{M}$ can handle much longer test motions than the training motions, we propose an autoregressive motion infilling process at test time as illustrated in Fig.~\ref{fig:mfiller} (Left). The key idea is to use a sliding window of $h$ frames, where we assume the first $h_\texttt{c}$ frames of motion are already occlusion-free or infilled and serve as \emph{context}, and we also use the last $h_\texttt{l}$ frames as \emph{look-ahead}. The look-ahead is essential to the motion infiller since it may contain visible poses that can guide the ending motion and avoid generating discontinuous motions. Excluding the context and look-ahead frames, only the middle $h_\texttt{o} = h - h_\texttt{c} - h_\texttt{l}$ frames of motion are infilled. We iteratively infill the motion using the sliding window and advance the window by $h_\texttt{o}$ frames every step.

\vspace{2mm}
\noindent\textbf{Motion Infiller Network.}
The overall network design of the CVAE-based motion infiller is outlined in Fig.~\ref{fig:mfiller}~(Right). In particular, we employ a Transformer-based seq2seq architecture, which consists of three parts: (1) a \emph{context network} that uses a Transformer encoder to encode the visible poses from the occluded motion $\mb{\wt{\Theta}}$ into a context sequence, which serves as the condition for other networks; (2) a \emph{decoder network} that uses the latent code $\mb{z}$ and context sequence to generate occlusion-free motion $\mb{\wh{\Theta}}$ via a Transformer decoder and a multilayer perceptron (MLP); (3) \emph{prior and posterior networks} that generate the prior and posterior distributions for the latent code $\mb{z}$. In the networks, we adopt a time-based encoding that replaces the position in the original positional encoding~\cite{vaswani2017attention} with the time index. Unlike prior CNN-based methods~\cite{hernandez2019human,kaufmann2020convolutional}, our Transformer-based motion infiller does not require padding missing frames, but instead restricts its attention to visible frames to achieve effective temporal modeling.

\vspace{2mm}
\noindent\textbf{Training.}
We train the motion infiller $\mathcal{M}$ using a large motion capture dataset, AMASS~\cite{AMASS:ICCV:2019}. To synthesize occluded motions $\mb{\wt{\Theta}}$, for any GT training motion $\mb{\wt{\Theta}}'$ of $h$ frames, we randomly occlude $H_\texttt{occ}$ consecutive frames of motion where $H_\texttt{occ}$ is uniformly sampled from $[H_\texttt{lb}, H_\texttt{ub}]$. Note that we do not occlude the first $h_\texttt{c}$ frames which are reserved as context. We use the standard CVAE objective to train the motion infiller $\mathcal{M}$:
\begin{align}
\label{eq:train_mfiller}
    L_\mathcal{M} = \sum_{t=1}^h \| \mb{\wt{\theta}}_t - \mb{\wt{\theta}}'_t \|_2^2 + L_\texttt{KL}^{\mb{z}}\,,
\end{align}
where $L_\texttt{KL}^{\mb{z}}$ is the KL divergence between the prior and posterior distributions of the CVAE latent code $\mb{z}$.

\subsection{Global Trajectory Predictor}
\label{sec:traj_pred}
After we obtain occlusion-free body motion $\mb{\wh{\Theta}}^i$ for each person using the motion infiller, a key problem still remains: the estimated trajectory $(\mb{\wt{T}}^i,\mb{\wt{R}}^i)$ of the person is still occluded and not in a consistent global coordinate system. To tackle this problem, we propose to learn a global trajectory predictor $\mathcal{T}$ that generates a person's occlusion-free global trajectory $(\mb{\wh{T}}^i, \mb{\wh{R}}^i)$ from the local body motion $\mb{\wh{\Theta}}^i$.

Given a general occlusion-free body motion ${\mb{\Theta}} = ({\mb{\theta}}_1, \ldots, {\mb{\theta}}_m)$ as input, the trajectory predictor $\mathcal{T}$ outputs its corresponding global trajectory $({\mb{T}}, {\mb{R}})$ including the root translations ${\mb{T}} = ({\mb{\tau}}_1, \ldots, {\mb{\tau}}_m)$ and rotations ${\mb{R}} = ({\mb{\rot}}_1, \ldots, {\mb{\rot}}_m)$. To address any potential ambiguity in the global trajectory, we also formulate the global trajectory predictor using the CVAE:
\begin{align}
\label{eq:traj_pred1}
\bs{{\Psi}} &= \mathcal{T} (\bs{{\Theta}}, \mb{v})\,, \\
\label{eq:traj_pred2}
(\mb{{T}}, \mb{{R}}) &= \texttt{EgoToGlobal}(\bs{{\Psi}})\,,
\end{align}
where the global trajectory predictor $\mathcal{T}$ corresponds to the CVAE decoder and $\mb{v}$ is the latent code for the CVAE. In Eq.~\eqref{eq:traj_pred1}, the immediate output of the global trajectory predictor $\mathcal{T}$ is an egocentric trajectory $\bs{{\Psi}} = (\bs{{\psi}}_1, \ldots, \bs{{\psi}}_m)$, which by design can be converted to a global trajectory $(\mb{{T}}, \mb{{R}})$ using a conversion function \texttt{EgoToGlobal}.

\vspace{2mm}
\noindent\textbf{Egocentric Trajectory Representation.}
The egocentric trajectory $\bs{{\Psi}}$ is just an alternative representation of the global trajectory $(\mb{{T}}, \mb{{R}})$. It converts the global trajectory into relative local differences and represents rotations and translations in the heading coordinates ($y$-axis aligned with the heading, \ie, the person's facing direction). In this way, the egocentric trajectory representation is invariant of the absolute $xy$ translation and heading. It is more suitable for the prediction of long trajectories, since the network only needs to output the local trajectory change of every frame instead of the potentially large global trajectory offset.

The conversion from the global trajectory to the egocentric trajectory is given by another function: $\bs{{\Psi}} = \texttt{GlobalToEgo}(\mb{{T}}, \mb{{R}})$, which is the inverse of the function \texttt{EgoToGlobal}. In particular, the egocentric trajectory $\bs{{\psi}}_t = (\delta x_t, \delta y_t, z_t, \delta \phi_t, \bs{\eta}_t)$ at time $t$ is computed as:
\begin{align}
\label{eq:ego_traj1}
(\delta x_t, \delta y_t) &= \texttt{ToHeading}(\mb{\tau}_t^{xy} - \mb{\tau}_{t-1}^{xy})\,,\\
\label{eq:ego_traj2}
z_t &= \mb{\tau}_t^z, \quad \delta \phi_t = \mb{\rot}_t^{\phi} - \mb{\rot}_{t-1}^{\phi}\,,\\
\label{eq:ego_traj3}
\bs{\eta}_t &= \texttt{ToHeading}(\mb{\rot}_t)\,,
\end{align}
where $\mb{\tau}_t^{xy}$ is the $xy$ component of the translation $\mb{\tau}_t$, $\mb{\tau}_t^{z}$ is the $z$ component (height) of $\mb{\tau}_t$, $\mb{\rot}_t^{\phi}$ is the heading angle of the rotation $\mb{\rot}_t$, \texttt{ToHeading} is a function that converts translations or rotations to the heading coordinates defined by the heading $\mb{\rot}_t^{\phi}$, and $\bs{\eta}_t$ is the local rotation. As an exception, $(\delta x_0, \delta y_0)$ and $\delta \phi_0$ are used to store the initial $xy$ translation $\mb{\tau}_0^{xy}$ and heading $\mb{\tau}_0^{\phi}$. These initial values are set to the GT during training and arbitrary values during inference (as the trajectory can start from any position and heading). The inverse process of Eq. \eqref{eq:ego_traj1}-\eqref{eq:ego_traj3} defines the inverse conversion \texttt{EgoToGlobal} used in Eq.~\eqref{eq:traj_pred2}, which accumulates the egocentric trajectory to obtain the global trajectory. To correct potential drifts in the trajectory, in Sec.~\ref{sec:global_opt}, we will optimize the global trajectory of each person to match the video evidence, which also solves the trajectory's starting point $(\delta x_0, \delta y_0, \delta \phi_0)$. More details about the egocentric trajectory are given in Appendix~\ref{sec:supp:traj_pred}.

\vspace{2mm}
\noindent\textbf{Network and Training.}
The trajectory predictor adopts a similar network design as the motion infiller with one main difference:  we use LSTMs for temporal modeling instead of Transformers since the output of each frame is the local trajectory change in our egocentric trajectory representation, which mainly depends on the body motion of nearby frames and does not require long-range temporal modeling. We will show in Sec.~\ref{sec:eval_component} that the egocentric trajectory and use of LSTMs instead of Transformers are crucial for accurate trajectory prediction. Please refer to Appendix~\ref{sec:supp:traj_pred} for the detailed network architectures. We use the standard CVAE objective to train the trajectory predictor $\mathcal{T}$:
\begin{align}
\label{eq:train_traj}
    L_\mathcal{T} = \sum_{t=1}^m \left(\| \mb{{\tau}}_t - \mb{{\tau}}'_t \|_2^2 + \| \mb{\rot}_t \ominus \mb{\rot}'_t \|_a^2\right) + L_\texttt{KL}^{\mb{v}}\,,
\end{align}
where $\mb{{\tau}}'_t$ and $\mb{\rot}'_t$ denote the GT translation and rotation, $\ominus$ computes the relative rotation, $\|\cdot\|_a$ computes the rotation angle, and $L_\texttt{KL}^{\mb{v}}$ is the KL divergence between the prior and posterior distributions of the CVAE latent code $\mb{v}$. We again use AMASS~\cite{AMASS:ICCV:2019} to train the trajectory predictor $\mathcal{T}$.

\subsection{Global Optimization}
\label{sec:global_opt}

After using the generative motion infiller and global trajectory predictor, we have obtained an occlusion-free global motion $\wh{\mb{Q}}^i = (\wh{\mb{T}}^i, \wh{\mb{R}}^i, \wh{\mb{\Theta}}^i, \wh{\mb{B}}^i)$ for each person in the video. However, the global trajectory predictor generates trajectories for each person independently, which may not be consistent with the video evidence. To tackle this problem, we propose a global optimization process that jointly optimizes the global trajectories of all people and the extrinsic camera parameters to match the video evidence such as 2D keypoints. The final output of the global optimization and our framework is $\wc{\mb{Q}}^i = (\wc{\mb{T}}^i, \wc{\mb{R}}^i, \wc{\mb{\Theta}}^i, \wc{\mb{B}}^i)$ where $(\wc{\mb{\Theta}}^i, \wc{\mb{B}}^i) = (\wh{\mb{\Theta}}^i, \wh{\mb{B}}^i)$, \ie, we directly use the occlusion-free body motion and shapes from the previous stages.

\vspace{3mm}
\noindent\textbf{Optimization Variables.}
The first set of variables we optimize is the egocentric representation $\{\wc{\mb{\Psi}}^i\}_{i=1}^N$ of the global trajectories $\{(\wc{\mb{T}}^i, \wc{\mb{R}}^i)\}_{i=1}^N$. We adopt the egocentric representation since it allows corrections of the translation and heading at one frame to propagate to all future frames. Therefore, it enables optimizing the trajectories of occluded frames since they will impact future visible frames under the egocentric trajectory representation. We will empirically demonstrate its effectiveness in Sec.~\ref{sec:eval_component}.

The second set of optimization variables is the extrinsic camera parameters $\mb{C} = (\mb{C}_1, \ldots, \mb{C}_T)$ where $\mb{C}_t \in \mathbb{R}^{4\times 4}$ is the camera extrinsic matrix at frame $t$ of the video.

\vspace{3mm}
\noindent\textbf{Energy Function.}
The energy function we aim to minimize is defined as
\begin{equation}
\label{eq:energy}
\begin{aligned}
    E(\{\wc{\mb{\Psi}}^i\}_{i=1}^N, \mb{C}) &= \lambda_\texttt{2D} E_\texttt{2D} + \lambda_\texttt{traj} E_\texttt{traj} \\
    & \hspace{-4mm} + \lambda_\texttt{reg} E_\texttt{reg} + \lambda_\texttt{cam} E_\texttt{cam} + \lambda_\texttt{pen} E_\texttt{pen}\,,
\end{aligned}
\end{equation}
where we use five energy terms with their corresponding coefficients $\lambda_\texttt{2D},\lambda_\texttt{traj},\lambda_\texttt{reg},\lambda_\texttt{cam},\lambda_\texttt{pen}$.

The first term $E_\texttt{2D}$ measures the error between  the 2D projection $\wc{\mb{x}}_t^i$ of the optimized 3D keypoints $\wc{\mb{X}}_t^i \in \mathbb{R}^{J \times 3}$ and the estimated 2D keypoints $\wt{\mb{x}}_t^i$ from a keypoint detector: 
\begin{align}
    E_\texttt{2D} = \frac{1}{NTJ}\sum_{i=1}^N & \sum_{t=1}^T V_t^i \|\wc{\mb{x}}_t^i - \wt{\mb{x}}_t^i\|_F^2\,, \\
    \wc{\mb{x}}_t^i = \Pi \left(\wc{\mb{X}}_t^i, \mb{C}_t, \mb{K} \right), &\quad \wc{\mb{X}}_t^i = \mathcal{J}(\wc{\mb{\tau}}_t^i, \wc{\mb{\rot}}_t^i, \wc{\mb{\theta}}_t^i, \wc{\mb{\beta}}_t^i)
\end{align}
where $V_t^i$ is person $i$'s visibility at frame $t$, $\Pi$ is the camera projection with extrinsics $\mb{C}_t$ and approximated intrinsics $\mb{K}$, and $\wc{\mb{X}}_t^i$ is computed using the SMPL joint function $\mathcal{J}$ from the optimized global pose $\wc{\mb{q}}_t^i = (\wc{\mb{\tau}}_t^i, \wc{\mb{\rot}}_t^i, \wc{\mb{\theta}}_t^i, \wc{\mb{\beta}}_t^i) \in \wc{\mb{Q}}^i$.

The second term $E_\texttt{traj}$ measures the difference between the optimized global trajectory $(\wc{\mb{T}}^i, \wc{\mb{R}}^i)$ viewed in the camera coordinates and the trajectory $(\wt{\mb{T}}^i, \wt{\mb{R}}^i)$ output by the pose estimator (\eg, KAMA~\cite{iqbal2021kama}) in Stage I:
\begin{equation}
\label{eq:E_traj}
\begin{aligned}
    E_\texttt{traj} = \frac{1}{NT}\sum_{i=1}^N\sum_{t=1}^T V_t^i &\left(\|\Gamma(\wc{\mb{\rot}}_t^i, \mb{C}_t) \ominus \wt{\mb{\rot}}_t^i\|_a^2 \right. \\
    + &\left. w_t \|\Gamma(\wc{\mb{\tau}}_t^i, \mb{C}_t) - \wt{\mb{\tau}}_t^i\|_2^2\right),
\end{aligned}
\end{equation}
where the function $\Gamma(\cdot, \mb{C}_t)$ transforms the global rotation $\wc{\mb{\rot}}_t^i$ or translation $\wc{\mb{\tau}}_t^i$ to the camera coordinates defined by $\mb{C}_t$, and $w_t$ is a weighting factor for the translation term.

The third term $E_\texttt{reg}$ regularizes the egocentric trajectory $\wc{\mb{\Psi}}^i$ to stay close to the output $\wh{\mb{\Psi}}^i$ of the trajectory predictor:
\begin{align}
\label{eq:E_reg}
    E_\texttt{reg} = \frac{1}{NT}\sum_{i=1}^N\sum_{t=1}^T \left\|\mb{w}_\psi\circ\left(\wc{\mb{\psi}}_t^i - \wh{\mb{\psi}}_t^i\right)\right\|_2^2,
\end{align}
where $\circ$ denotes the element-wise product and $\mb{w}_\psi$ is a weighting vector for each element inside the egocentric trajectory. As an exception, we do not regularize each person's initial $xy$ position and heading $(\delta \wc{x}^i_0, \delta \wc{y}^i_0, \delta \wc{\phi}^i_0) \subset \wc{\mb{\psi}}_0^i$ as they need to be inferred from the video.

The fourth term $E_\texttt{cam}$ measures the smoothness of the camera parameters $\mb{C}$ and the uprightness of the camera:
\begin{equation}
\begin{aligned}
    E_\texttt{cam} & = \frac{1}{T}\sum_{t=1}^{T} \langle \mb{C}_t^y, \mb{Y} \rangle \\
    & \hspace{-5mm} +\frac{1}{T-1}\sum_{t=1}^{T-1} \left\|\mb{C}_{t+1}^\rot \ominus \mb{C}_t^\rot\right\|_a^2 + \left\|\mb{C}_{t+1}^\tau - \mb{C}_t^\tau\right\|_2^2,
\end{aligned}
\end{equation}
where $\langle\cdot,\cdot\rangle$ denotes the inner product, $\mb{C}_t^y$ is the $+y$ vector of the camera $\mb{C}_t$, and $\mb{Y}$ is the global up direction. $\mb{C}_t^\rot$ and $\mb{C}_t^\tau$ denote the rotation and translation of the camera $\mb{C}_t$.

The final term $E_\texttt{pen}$ is an signed distance field (SDF)-based inter-person penetration loss adopted from~\cite{jiang2020coherent}.

\section{Experiments}
\label{sec:exp}

\noindent\textbf{Datasets.}
We employ the following datasets in our experiments:
(1) \textbf{AMASS}~\cite{AMASS:ICCV:2019}, which is a large human motion database with 11000+ human motions. We use AMASS to train and evaluate the motion infiller and trajectory predictor.
(2)~\textbf{3DPW}~\cite{vonMarcard2018}, which is an \emph{in-the-wild} human motion dataset that uses videos and wearable IMU sensors to obtain GT poses, even when the person is occluded. We evaluate our approach using the test split of 3DPW.
(3)~\textbf{\mbox{Dynamic Human3.6M}} is a new benchmark for human pose estimation with dynamic cameras that we create from the Human3.6M dataset~\cite{h36m_pami}. We simulate dynamic cameras and occlusions by cropping each frame with a small view window that oscillates around the person (see Fig.~\ref{fig:results_vis_h36m}). More details are provided in Appendix~\ref{sec:supp:dataset}.

\vspace{2mm}
\noindent\textbf{Evaluation Metrics.}
We use the following metrics for evaluation: (1) \textbf{G-MPJPE} and \textbf{G-PVE}, which extend the mean per joint position error (MPJPE) and per-vertex error (PVE) by computing the errors in the global coordinates. As errors in estimated global trajectories accumulate over time in our dynamic camera setting, we follow standard evaluations for open-loop reconstruction (\eg, SLAM~\cite{sturm2011towards} and inertial odometry~\cite{herath2020ronin}) to compute errors using a sliding window (10 seconds) and align the root translation and rotation with the GT at the start of the window. (2) \textbf{PA-MPJPE}, which is the Procrustes-aligned MPJPE for evaluating estimated body poses. For invisible poses, since there can be many plausible poses beside the GT, we follow prior work~\cite{aliakbarian2020stochastic,yuan2020dlow} to compute the best PA-MPJPE out of multiple samples for our probabilistic approach. (3) \textbf{Accel}, which computes the mean acceleration error of each joint and is commonly used to measure the jitter in estimated motions~\cite{yuan2021simpoe,kocabas2020vibe}. (4)~\textbf{FID}, which is an extension of the original Frechet Inception Distance that calculates the distribution distance between estimated motions and the GT. FID is a standard metric in motion generation literature to evaluate the quality of generated motions~\cite{li2020learning,valle2021transflower,huang2021,li2021ai}. Following prior work~\cite{li2021ai}, we compute FID using the well-designed kinetic motion feature extractor in the fairmotion library~\cite{gopinath2020fairmotion}.

\vspace{2mm}
\noindent\textbf{Implementation Details.}
Thorough details about the entire framework are provided in Appendix~\ref{sec:supp:dataset} to \ref{sec:supp:global_opt}.

\begin{figure*}[t]
\centering
\includegraphics[width=\linewidth]{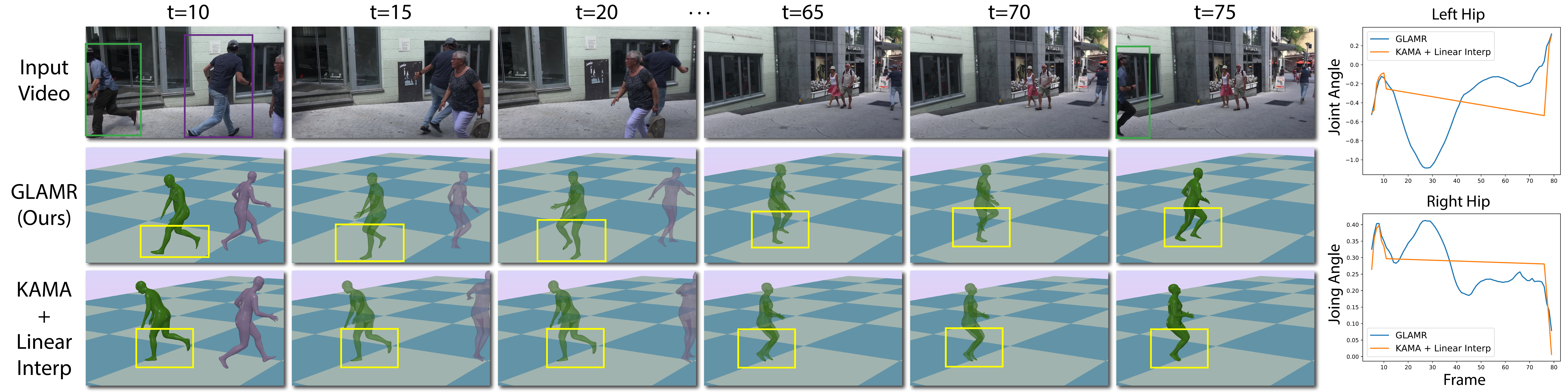}
\vspace{-6mm}
\caption{Qualitative comparison of GLAMR with a strong baseline on 3DPW. The infilled motion (transparent) by GLAMR is more natural especially for the legs, while the baseline has very slow leg motions due to interpolation in a large window (frame 10 to 75).  On the \textbf{right}, we plot how the $x$-axis joint angles of left and right hips of the person (green) change over time for GLAMR and the baseline.}
\label{fig:results_vis}
\vspace{-4mm}
\end{figure*}

\subsection{Evaluation of GLAMR}
\label{sec:eval_glamr}
\noindent\textbf{Baselines.}
Since no prior methods can estimate global motions from dynamic cameras and address long-term occlusions, we design various baselines by combining state-of-the-art human mesh recovery methods (KAMA~\cite{iqbal2021kama} or SPEC~\cite{Kocabas_SPEC_2021}), motion infilling methods, and SLAM-based camera estimation (OpenSfM~\cite{opensfm2021}). In particular, we use the estimated camera parameters to convert estimated motions from the camera coordinates to the global coordinates.  For motion infilling, we use (1) linear interpolation, (2) last pose, \ie, replicating the last visible pose, and (3) a state-of-the-art CNN-based motion infilling method, ConvAE~\cite{kaufmann2020convolutional}. 

The results on Dynamic Human3.6M and 3DPW are summarized in Table~\ref{table:baseline_h36m} and \ref{table:baseline_3dpw} respectively. We only report G-MPJPE and G-PVE on Dynamic Human3.6M since they require accurate GT trajectories, which 3DPW does not provide. It is evident that our approach, GLAMR, outperforms the baselines in almost all metrics. In particular, GLAMR achieves significantly lower G-MPJPE and G-PVE, which demonstrates its strong ability to reconstruct global human motions. Furthermore, GLAMR attains considerably lower FID and PA-MPJPE (with ten samples) for occluded (invisible) poses. The lower FID means GLAMR can infill more humanlike motions, and the lower PA-MPJPE also shows GLAMR's probabilistic motion samples can cover the GT better. Finally, while GLAMR achieves almost the same PA-MPJPE for visible poses as the best method, it yields much smoother motions (smaller acceleration error). This is because our motion infiller leverages human dynamics learned from a large motion dataset to produce motions.

\setlength{\tabcolsep}{3pt}
\begin{table}[t]
\footnotesize
\centering
\resizebox{\linewidth}{!}{
\begin{tabular}{@{\hskip 1mm}lccccccc@{\hskip 1mm}}
\toprule
Method & \begin{tabular}{@{}c@{}}(All)  \\ G-MPJPE \end{tabular} & \begin{tabular}{@{}c@{}}(All)  \\ G-PVE \end{tabular} & \begin{tabular}{@{}c@{}}(Invisible)  \\ FID \end{tabular} & \begin{tabular}{@{}c@{}}(Invisible)  \\ PA-MPJPE \end{tabular} & \begin{tabular}{@{}c@{}}(Visible) \\ PA-MPJPE \end{tabular} & \begin{tabular}{@{}c@{}}(All)  \\ Accel \end{tabular} \\\midrule
KAMA~\cite{kaufmann2020convolutional} + Linear Interpolation & 1735.2 & 1744.1 & 30.2 & 74.8 & \textbf{47.4} & 8.0 \\
KAMA~\cite{kaufmann2020convolutional} + Last Pose & 1318.1 & 1330.3 & 36.7 & 88.8 & \textbf{47.4} & 12.3 \\
KAMA~\cite{kaufmann2020convolutional} + ConvAE~\cite{kaufmann2020convolutional} & 1737.8 & 1748.9 & 28.9 & 77.4 & 56.9 & 7.5 \\
SPEC~\cite{Kocabas_SPEC_2021} + Linear Interpolation & 2113.3 & 2119.5 & 29.7 & 78.7 & 55.7 & 14.2 \\
SPEC~\cite{Kocabas_SPEC_2021} + Last Pose & 1782.5 & 1790.9 & 36.2 & 92.6 & 55.7 & 18.8 \\
SPEC~\cite{Kocabas_SPEC_2021} + ConvAE~\cite{kaufmann2020convolutional} & 2113.3 & 2119.0 & 28.5 & 80.1 & 59.9 & 11.9 \\ \midrule
Ours (GLAMR w/ SPEC) & 899.1 & 913.7 & \textbf{8.2} & 72.8 & 55.0 & 6.6 \\
Ours (GLAMR w/ KAMA) & \textbf{806.2} & \textbf{824.1} & 11.4 & \textbf{67.7} & 47.6 & \textbf{6.0} \\
\bottomrule
\end{tabular}
}
\vspace{-2.5mm}
\caption{Baseline comparison on Dynamic Human3.6M. We report results for visible, invisible (occluded), and all frames.}
\label{table:baseline_h36m}
\vspace{-3mm}
\end{table}

\setlength{\tabcolsep}{5pt}
\begin{table}[t]
\footnotesize
\centering
\resizebox{\linewidth}{!}{ 
\begin{tabular}{@{\hskip 1mm}lccccc@{\hskip 1mm}}
\toprule
Method & \begin{tabular}{@{}c@{}}(Invisible)  \\ FID \end{tabular} & \begin{tabular}{@{}c@{}}(Invisible)  \\ PA-MPJPE \end{tabular} & \begin{tabular}{@{}c@{}}(Visible) \\ PA-MPJPE \end{tabular} & \begin{tabular}{@{}c@{}}(All)  \\ Accel \end{tabular} \\ \midrule
KAMA~\cite{kaufmann2020convolutional} + Linear Interpolation & 30.7 & 87.5 & \textbf{50.8} & 24.2 \\
KAMA~\cite{kaufmann2020convolutional} + Last Pose & 40.3 & 96.3 & \textbf{50.8} & 25.4 \\
KAMA~\cite{kaufmann2020convolutional} + ConvAE~\cite{kaufmann2020convolutional} & 32.0 & 84.5 & 56.4 & 19.6 \\
SPEC~\cite{Kocabas_SPEC_2021} + Linear Interpolation & 33.6 & 85.6 & 53.3 & 33.1 \\
SPEC~\cite{Kocabas_SPEC_2021} + Last Pose & 39.5 & 92.4 & 53.3 & 34.2 \\
SPEC~\cite{Kocabas_SPEC_2021} + ConvAE~\cite{kaufmann2020convolutional} & 35.4 & 86.9 & 59.3 & 24.0 \\ \midrule
Ours (GLAMR w/ SPEC) & 24.8 & 79.1 & 54.9 & 9.5 \\
Ours (GLAMR w/ KAMA) & \textbf{22.6} & \textbf{73.6} & 51.1 & \textbf{8.9} \\
\bottomrule
\end{tabular}
}
\vspace{-2mm}
\caption{Baseline comparison on 3DPW. G-MPJPE and G-PVE are not reported since 3DPW does not provide accurate GT global human trajectories. See also the caption of Table~\ref{table:baseline_h36m}.}
\label{table:baseline_3dpw}
\vspace{-3.5mm}
\end{table}

\begin{figure}[t]
\vspace{-1mm}
\centering
\includegraphics[width=\linewidth]{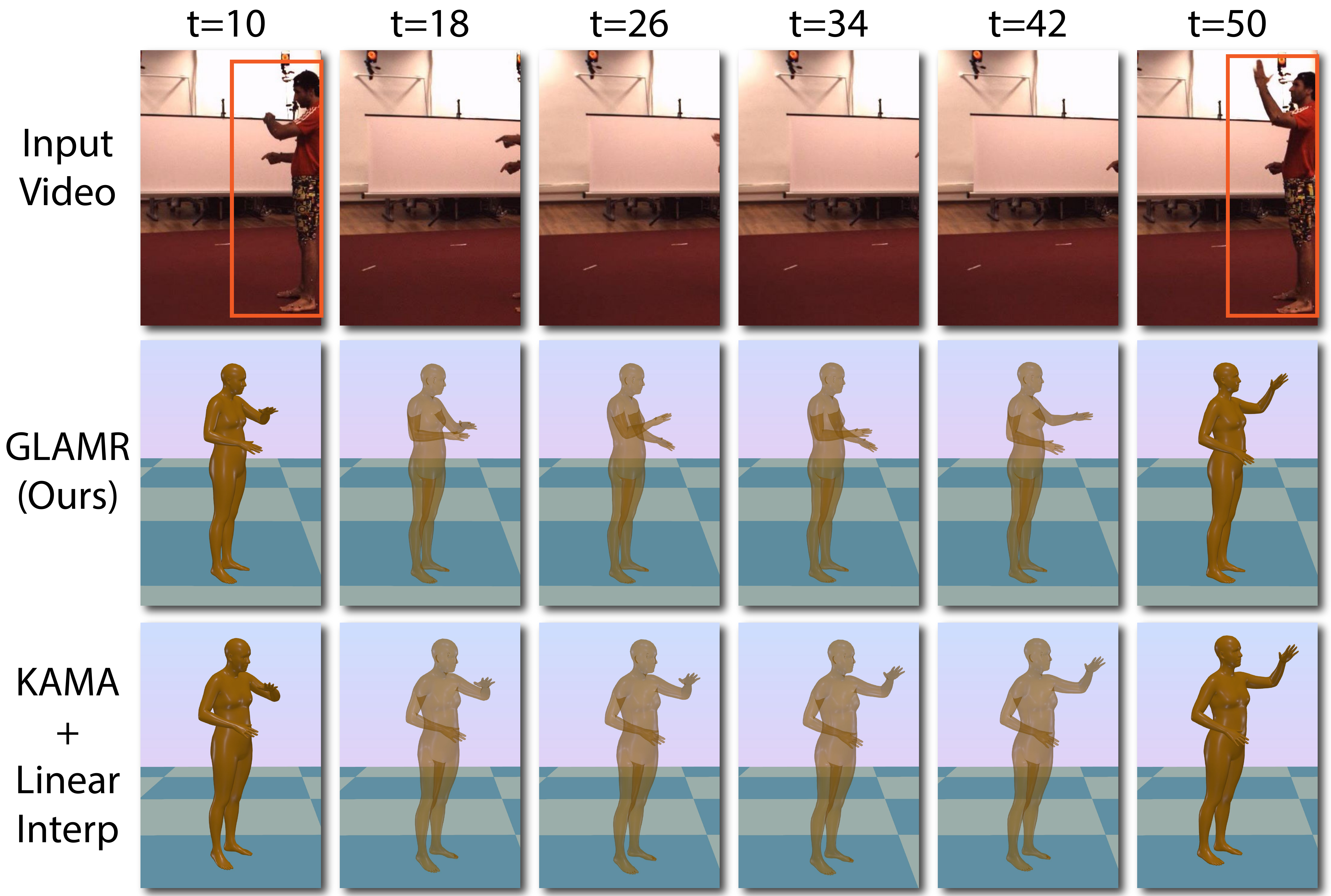}
\vspace{-6mm}
\caption{Qualitative comparison of GLAMR on Dynamic Human3.6M.  GLAMR can generate natural hand motions for invisible frames instead of just doing linear interpolation.}
\label{fig:results_vis_h36m}
\vspace{-5mm}
\end{figure}

\vspace{2mm}
\noindent\textbf{Qualitative Results.}
Fig.~\ref{fig:results_vis} and \ref{fig:results_vis_h36m} show qualitative comparisons of GLAMR against the strong baseline, KAMA + Linear Interpolation. Additionally, we provide abundant qualitative results on the \href{https://nvlabs.github.io/GLAMR}{project page}.

\subsection{Evaluation of Key Components}
\label{sec:eval_component}

\noindent\textbf{Benchmarking Motion Infiller.}
We evaluate the proposed generative motion infiller on the test split of the AMASS dataset~\cite{AMASS:ICCV:2019}. We compare against three motion infilling baselines: linear interpolation, replicating the last pose, and \mbox{ConvAE}~\cite{kaufmann2020convolutional}. As shown in Table~\ref{table:mfiller}, our generative motion infiller achieves significantly better PA-MPJPE for both the sampled motions (with five samples) and reconstructed motion for the infilled frames. Our approach also achieves considerably better FID, reducing the FID of ConvAE~\cite{kaufmann2020convolutional} by half, which indicates that the infilled motions by our approach are much closer to real human motions.

\vspace{1.25mm}
\noindent\textbf{Benchmarking Trajectory Predictor.}
We also evaluate our global trajectory predictor against two variants on the AMASS test set: (1) ``Transformer'', which replaces the LSTMs in the trajectory predictor with Transformers; (2)~``Ours w/o Ego Trajectory'', which does not use the egocentric trajectory but instead directly outputs the 6-DoF global trajectory. As shown in Table~\ref{table:traj_pred}, both variants lead to worse global trajectory prediction (higher best-of-five G-MPJPE and G-PVE). We believe the reasons are: (1) the positional encoding in Transformers may not generalize well to longer motions compared to the LSTMs in our approach; (2) directly predicting the 6-DoF global trajectory offsets instead of egocentric trajectories from local body motions is also hard to generalize since the global offsets can be large.

\vspace{1.25mm}
\noindent\textbf{Ablations for Global Optimization.}
We further perform ablation studies on the effect of key components in our global optimization. Specifically, we design two variants: (1) ``Ours w/o Trajectory Predictor'', which does not use our trajectory predictor to generate the global human trajectories and uses camera parameters from OpenSfM~\cite{opensfm2021} to obtain global trajectories instead; (2) ``Ours w/o Opt Ego Trajectory'', which does not employ the egocentric trajectory representation and directly optimizes the 6-DoF root trajectory instead. As shown in Table~\ref{table:global_opt}, both variants lead to significantly worse global trajectory reconstruction with large increases in G-MPJPE, G-PVE, and Accel. This demonstrates that both the global trajectory predictor and egocentric trajectory representation are vital in our approach.

\setlength{\tabcolsep}{5pt}
\begin{table}[t]
\footnotesize
\centering
\resizebox{0.9\linewidth}{!}{ 
\begin{tabular}{@{\hskip 1mm}lccccc@{\hskip 1mm}}
\toprule
Method & \begin{tabular}{@{}c@{}}(Sampled) \\ PA-MPJPE \end{tabular} & \begin{tabular}{@{}c@{}}(Reconstructed) \\ PA-MPJPE \end{tabular} & \begin{tabular}{@{}c@{}}(Sampled) \\ FID \end{tabular}\\ \midrule
Linear Interpolation & 83.5 & 83.5 & 35.3 \\
Last Pose & 104.4 & 104.4 & 41.6 \\
ConvAE~\cite{kaufmann2020convolutional} & 72.8	& 72.8 & 31.4 \\
Ours & \textbf{61.4} & \textbf{36.1} & \textbf{16.7}\\
\bottomrule
\end{tabular}
}
\vspace{-2.5mm}
\caption{Benchmarking motion infiller on AMASS.}
\label{table:mfiller}
\vspace{-3mm}
\end{table}

\setlength{\tabcolsep}{5pt}
\begin{table}[t]
\footnotesize
\centering
\resizebox{0.9\linewidth}{!}{ 
\begin{tabular}{@{\hskip 1mm}lcccc@{\hskip 1mm}}
\toprule
Method & G-MPJPE & G-PVE & Accel \\ \midrule
Transformer & 660.1 & 678.6 & 121.9 \\
Ours w/o Ego Trajectory &  763.0 & 780.6 & 8.7 \\
Ours & \textbf{466.9} & \textbf{472.5} & \textbf{5.8} \\
\bottomrule
\end{tabular}
}
\vspace{-2.5mm}
\caption{Benchmarking trajectory predictor on AMASS.}
\label{table:traj_pred}
\vspace{-3mm}
\end{table}

\setlength{\tabcolsep}{5pt}
\begin{table}[t]
\footnotesize
\centering
\resizebox{0.9\linewidth}{!}{ 
\begin{tabular}{@{\hskip 1mm}lcccc@{\hskip 1mm}}
\toprule
Method & G-MPJPE & G-PVE & Accel \\ \midrule
Ours w/o Trajectory Predictor & 1750.8 & 1761.4 & 12.6 \\
Ours w/o Opt Ego Trajectory &  877.3 & 895.0 & 15.5 \\
Ours (GLAMR) & \textbf{806.2} & \textbf{824.1} & \textbf{6.0} \\
\bottomrule
\end{tabular}
}
\vspace{-2.5mm}
\caption{Global optimization ablations on Dynamic Human3.6M.}
\label{table:global_opt}
\vspace{-4mm}
\end{table}

\section{Discussion and Limitations}
In this paper, we proposed an approach for 3D human mesh recovery in consistent global coordinates from videos captured by dynamic cameras. We first proposed a novel Transformer-based generative motion infiller to address severe occlusions that often come with dynamic cameras. To resolve ambiguity in the joint reconstruction of global human motions and camera poses, we proposed a new solution by predicting global human trajectories from local body motions. Finally, we proposed a global optimization framework to refine the predicted trajectories, which serve as anchors for camera optimization. Our method achieves SOTA results on challenging datasets and marks a significant step towards global human mesh recovery in the wild.

As the first paper on this new problem, our method has a few limitations: propagation of errors in multiple stages, limited body shape estimation, not being real-time, not including scene information, \etc. A detailed discussion is provided in Appendix~\ref{sec:supp:limitations}. We believe these limitations are exciting avenues for future work to explore.

{\small
\bibliographystyle{ieee_fullname}
\bibliography{reference}
}

\appendix
\onecolumn

\section{Details for the Datasets}
\label{sec:supp:dataset}

\noindent\textbf{AMASS}~\cite{AMASS:ICCV:2019} is a large human motion database with 11000+ human motions. We use AMASS to train and evaluate the motion infiller and trajectory predictor. Specifically, we use the Transitions, SSM, and HumanEva~\cite{sigal2006humaneva} subsets for testing and all other subsets for training.

\vspace{1mm}
\noindent\textbf{3DPW}~\cite{vonMarcard2018} is an in-the-wild human motion dataset that consists of 60 videos recorded with dynamic cameras in diverse environments. The GT 3D poses are obtained using wearable IMU sensors. Since non-optical sensors are used to obtain GT data, the dataset also provides body pose information when the persons go outside the FoV of the camera. 3DPW also provides the global trajectories of people in the dataset. However, the global trajectories are quite inaccurate since they are estimated from IMU data. Therefore, we do not use 3DPW to evaluate global trajectory reconstruction in the paper. Since we do not use 3DPW for training, we use sequences from the entire 3DPW dataset for visualization. We use the official 3DPW test split to report quantitative results in the paper.

\vspace{1mm}
\noindent\textbf{\mbox{Dynamic Human3.6M}} is a new benchmark for global human pose estimation with dynamic cameras that we create from the Human3.6M dataset~\cite{h36m_pami}. We simulate dynamic cameras and occlusions by cropping each frame with a view window of $300\times 600$ that horizontally oscillates around the person's bounding box center with a period of 4.8 seconds and a magnitude of 200 pixels. In this way, we synthesize large camera motions and severe occlusions where the person is occluded for almost half of the time, which makes it very challenging for existing 3D human pose and shape estimation methods. Additionally, since Human3.6M provides accurate global human trajectories and human poses, we use Dynamic Human3.6M to evaluate global trajectory reconstruction and pose estimation for occluded frames. We follow the standard protocol~\cite{hmrKanazawa18} and use the official test split (subjects 9 and 11) for evaluation. Please refer to the [supplementary video](https://youtu.be/wpObDXcYueo) for an example sequence of the Dynamic Human3.6M dataset. Code for generating Dynamic Human3.6M are available \href{https://github.com/NVlabs/GLAMR}{here} for users who have downloaded the original Human3.6M dataset~\cite{h36m_pami}.

\section{Implementation Details for Preprocessing}
\label{sec:supp:preprocess}

\noindent\textbf{3D Multi-Object Tracking and Re-identification.}
We use DeepSORT~\cite{wojke2017simple} with ResNet-50~\cite{he2016deep} in the MMTracking package~\cite{mmtrack2020} for 3D multi-object tracking (MOT) and re-identification. We use the GT tracks to evaluate our approach and the baselines, following the standard protocol for human pose estimation.

\vspace{1mm}
\noindent\textbf{Initial Human Pose and Shape Estimation.}
As mentioned in the main paper, we use KAMA~\cite{iqbal2021kama} or SPEC~\cite{Kocabas_SPEC_2021} to provide the initial human pose and shape estimation from the bounding boxes extracted by 3D MOT. We choose these two methods since both KAMA and SPEC estimate 3D human poses in the camera coordinates with absolute root translations, while many state-of-the-art human pose estimation methods do not provide the root translations. We also use HRNet~\cite{sun2019deep} to extract 2D human keypoints from the video, which are used in the proposed global optimization framework.

\section{Implementation Details for Generative Motion Infiller}
\label{sec:supp:mfiller}

\begin{figure*}[ht]
    \vspace{-5mm}
    \centering
    \includegraphics[width=\linewidth]{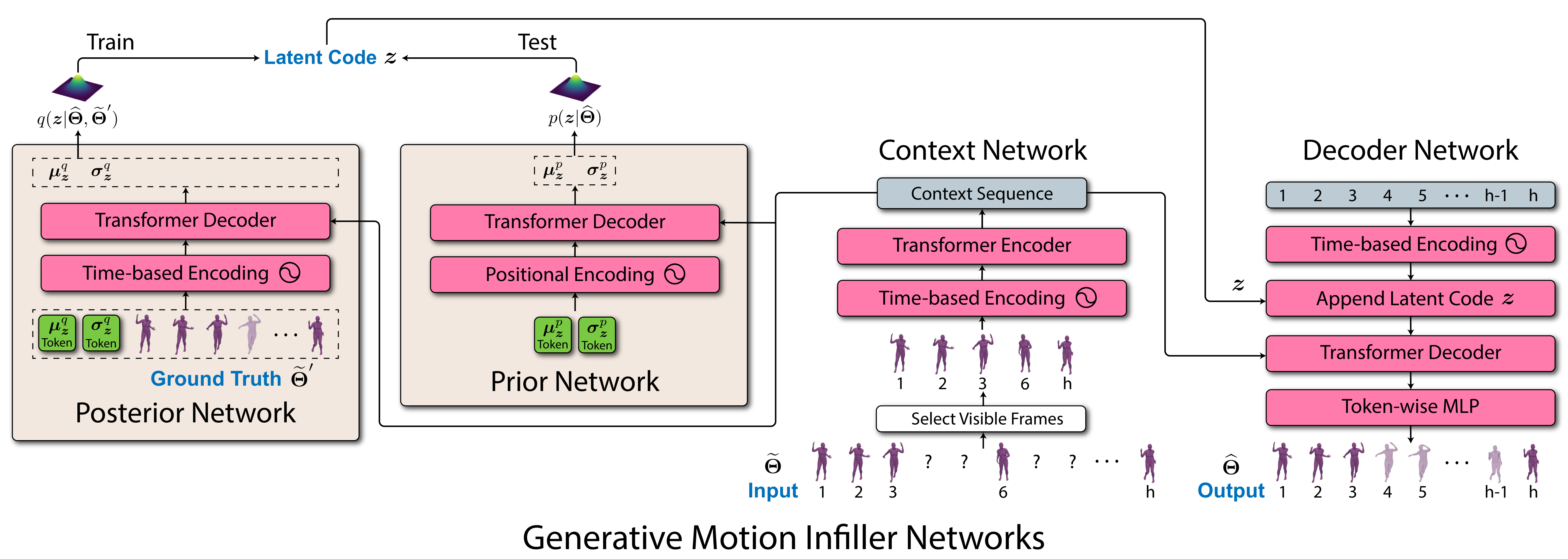}
    \vspace{-6mm}
    \caption{The detailed network architecture of the CVAE-based generative motion infiller. For all the Transformer modules, the dimensions for keys, queries, and values are set to 256, the number of transformer blocks is 2, the hidden dimensions of the feedforwards layers are 512, the dropout rate is 0.1, and 8 heads are used for the multi-head attention. Two hidden layers (512, 256) with ReLU activations are used for all the token-wise MLPs.}
    \label{fig:mfiller_supp}
    \vspace{-3mm}
\end{figure*}

\noindent\textbf{Network Architecture.}
The detailed network architecture of the CVAE-based generative motion infiller is outlined in Fig.~\ref{fig:mfiller_supp}. For all the Transformer~\cite{vaswani2017attention} modules, the dimensions for keys, queries, and values are set to 256, the number of transformer blocks is 2, the hidden dimensions of the feedforwards layers are 512, the dropout rate is 0.1, and 8 heads are used for the multi-head attention. The time-based encoding takes the same sinusoidal form as the original positional encoding~\cite{vaswani2017attention} but replaces the position with the time index. We use two hidden layers (512, 256) with ReLU activations for all the token-wise MLPs. In the prior network, two learnable tokens are used to form queries to produce the mean $\bs{\mu}_{\bs{z}}^p$ and standard deviation $\bs{\sigma}_{\bs{z}}^p$ of the prior distribution of the latent code $\bs{z}$. Similarly, in the posterior network, two learnable tokens are appended to the GT pose sequence $\tilde{\mb{\Theta}}'$ to output the mean $\bs{\mu}_{\bs{z}}^q$ and standard deviation $\bs{\sigma}_{\bs{z}}^q$ of the posterior distribution of the latent code~$\bs{z}$.

\vspace{2mm}
\noindent\textbf{Hyperparameters and Training.}
The dimension of the latent code $\mb{z}$ is 128. The sliding window size $h$ of the autoregressive motion infilling is 50. Both the number of context frames $h_\texttt{c}$ and the number of look-ahead $h_\texttt{l}$ frames are 10. When synthesizing occluded motions, for any GT training motion of $h=50$ frames, we randomly occlude $H_\texttt{occ}$ consecutive frames of motion where $H_\texttt{occ}$ is uniformly sampled from $[10, 40]$. Note that we do not occlude the first $h_\texttt{c}=10$ frames which are reserved as context. The KL divergence term in Eq.~\eqref{eq:train_mfiller} uses a weighting factor of 0.001. We train the networks for 2000 epochs with a batch size of 1024 where each epoch uses a total of 10 million frames of motion. For optimization, we use the Adam optimizer~\cite{kingma2014adam} with a learning rate of 0.001 and clip the gradient if its norm is larger than 5. We use PyTorch~\cite{paszke2019pytorch} to implement and train the networks.

\section{Implementation Details for Global Trajectory Predictor}
\label{sec:supp:traj_pred}

\noindent\textbf{Heading Coordinate and Egocentric Trajectory Representation.} The heading vector of a person points towards where the person is facing and is parallel to the ground. We obtain the heading vector by aligning the $z$-axis of the person's root coordinate with the world $z$-axis and use the resulting $y$-axis of the aligned root coordinate as the heading vector. This way of obtaining the heading is more stable than using the yaw of the Euler angle representation, which suffers from singularities and can be quite unstable. The heading coordinate is defined by first placing the world coordinate at the root position of the person and then rotating the world coordinate around the $z$-axis (vertical) to align the $y$-axis with the heading vector. By definition, representing and predicting human trajectories in the heading coordinate allows the predicted trajectory to be invariant of the person's absolute $xy$ translation and heading. In the egocentric trajectory representation $\bs{{\psi}}_t = (\delta x_t, \delta y_t, z_t, \delta \phi_t, \bs{\eta}_t)$, we use absolute height $z_t$ since the height of a person relative to the ground does not vary a lot and is highly correlated with the body motion of the person. For the local rotation $\bs{\eta}_t$, we adopt the 6D rotation representation \cite{zhou2019continuity} to avoid discontinuity.

\begin{figure*}[h]
    \centering
    \includegraphics[width=\linewidth]{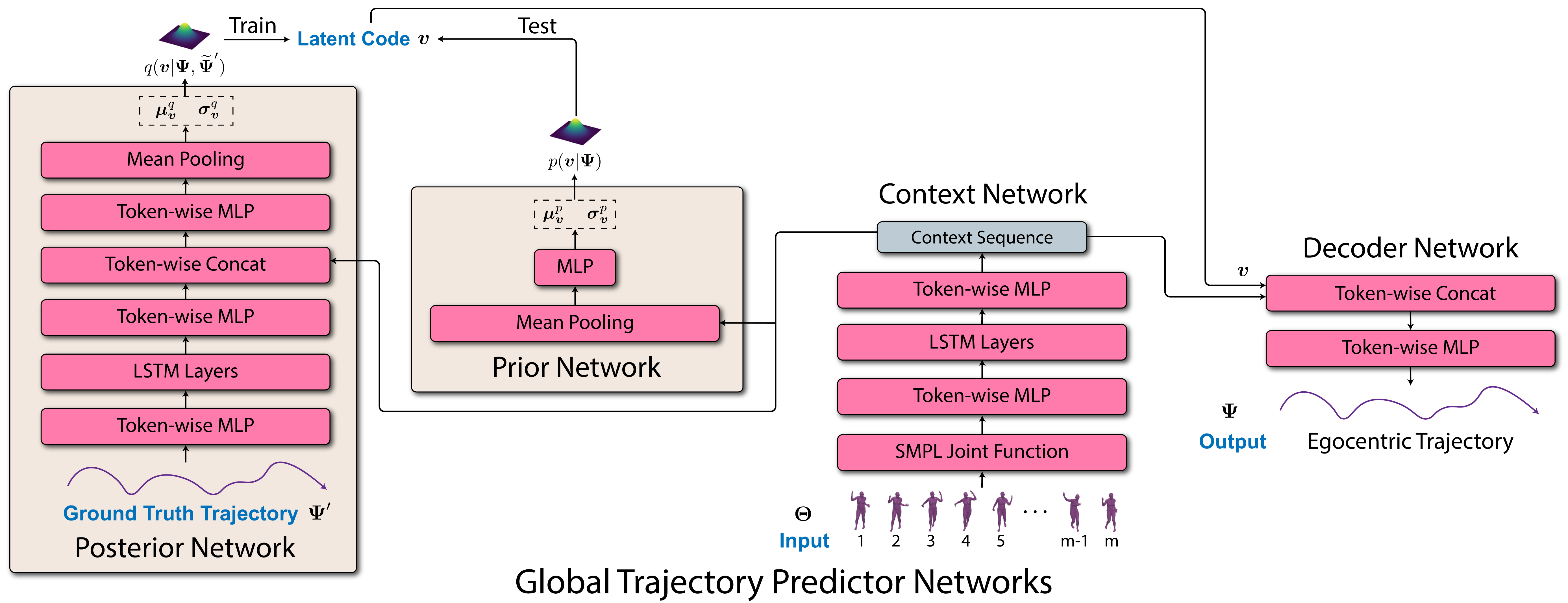}
    \vspace{-6mm}
    \caption{The network architecture of the CVAE-based global trajectory predictor. We use two bidirectional LSTM layers with hidden dimension 256 for all the LSTM blocks, and we use two hidden layers (512, 256) with ReLU activations for all the token-wise MLPs. Token-wise mean pooling is used in the prior and posterior networks to summary sequences into a single feature.}
    \label{fig:traj_pred_supp}
    \vspace{-3mm}
\end{figure*}

\vspace{2mm}
\noindent\textbf{Network Architecture.}
The detailed network architecture of the CVAE-based global trajectory predictor is illustrated in Fig.~\ref{fig:traj_pred_supp}. We use two bidirectional LSTM layers with hidden dimension 256 for all the LSTM blocks in the networks. We use two hidden layers (512, 256) with ReLU activations for all the token-wise MLPs. For the input poses, we first convert them to 3D joint positions using the SMPL joint function without global rotations and translations. This is because we find that using 3D joint positions leads to better performance than using joint rotations directly. In both the prior and posterior networks, token-wise mean pooling is used to produce a single feature from a sequence of tokens, which is then used to produce the parameters of the prior or posterior distribution of the latent code $\bs{v}$.

\vspace{2mm}
\noindent\textbf{Hyperparameters and Training.}
The dimension of the latent code $\mb{v}$ is 128. The KL divergence term in Eq.~\eqref{eq:train_traj} uses a weighting factor of 0.001. We train the networks for 2000 epochs with a batch size of 256 where each epoch uses a total of 2 million frames of motion. The training sequence length is 100 frames For optimization, we use the Adam optimizer~\cite{kingma2014adam} with a learning rate of 0.0001 and clip the gradient if its norm is larger than 5.  We use PyTorch~\cite{paszke2019pytorch} to implement and train the networks.

\section{Implementation Details for Global Optimization}
\label{sec:supp:global_opt}

\noindent\textbf{Initialization.}
We initialize the egocentric trajectories using the output from the global trajectory predictor. For the camera, we approximate the camera intrinsic parameters $\bs{K}$ using the dimensions of the image $[\texttt{w}, \texttt{h}]$ where we assume the principal point is at the image center $[\texttt{w}/2, \texttt{h}/2]$. Note that the camera intrinsics are kept fixed during the optimization process. For the camera extrinsic parameters $\bs{C}$, we initialize them from the persons' global trajectories using the following equations:
\begin{equation}
\label{eq:cam_init}
    \bs{C}_t = \Omega\left(\frac{1}{\sum_{i=1}^N V_t^i}\sum_{i=1}^N V_t^i \cdot \bs{P}_t^{i,\texttt{global}} {\bs{P}_t^{i,\texttt{cam}}}^{-1}\right)\,,
\end{equation}
where $V_t^i$ is the visibility of person $i$ at frame $t$,  $\bs{P}_t^{i,\texttt{global}} \in \mathbb{R}^{4\times 4}$ is the person's transformation in the global coordinates based on the predicted global trajectory $(\wh{\mb{T}}^i, \wh{\mb{R}}^i)$, $\bs{P}_t^{i,\texttt{cam}}\in \mathbb{R}^{4\times 4}$ is the person's transformation in the camera coordinates based on the estimated trajectory $(\wt{\mb{T}}^i, \wt{\mb{R}}^i)$ by the pose estimator (e.g., KAMA~\cite{iqbal2021kama}), $\Omega$ is a projection operator that projects the matrix into a valid transformation. If no person is visible at frame $t$, the camera extrinsics $\bs{C}_t$ is initialized to the camera extrinsics of the most recent frame with visible people. Eq.~\eqref{eq:cam_init} is the least squares solutions of the following (transposed) linear systems:
\begin{equation}
    \bs{P}_t^{i,\texttt{global}} = \bs{C}_t \bs{P}_t^{i,\texttt{cam}}\,, \quad\quad \forall i, V_t^i = 1\,.
\end{equation}

\vspace{2mm}
\noindent\textbf{Hyperparameters and Optimization.}
The optimization loss coefficients $(\lambda_\texttt{2D},\lambda_\texttt{traj},\lambda_\texttt{reg},\lambda_\texttt{cam},\lambda_\texttt{pen})$ in Eq.~\eqref{eq:energy} are set to (1, 100000, 100, 10000, 100000) for 3DPW and (1, 100000, 100, 10000, 0) for Human3.6M. We do not use the inter-person penetration loss for Human3.6M since it only has one person in each video. The weighting factor $w_t$ for the translation term in Eq.~\eqref{eq:E_traj} is set to 0 since the translation estimated by the pose estimator can be quite noisy. The trajectory regularization weighting factor $\mb{w}_\psi$ in Eq.~\eqref{eq:E_reg} is set to (3,10,10000,5,10000) for each element in the egocentric trajectory $\bs{{\psi}}_t = (\delta x_t, \delta y_t, z_t, \delta \phi_t, \bs{\eta}_t)$, where we use large weights to penalize changes in height $z_t$ and local rotation $\bs{\eta}_t$. The global optimization is also implemented in PyTorch~\cite{paszke2019pytorch}, where we use the Adam optimizer~\cite{kingma2014adam} with a learning rate of 0.001 to optimize the global trajectories and camera extrinsics.

\vspace{2mm}
\noindent\textbf{Computation Time.}
The overall processing time for a 1-min scene is around 5 mins with 500 optimization iterations, which is much faster than using OpenSfM ($>30$ mins).

\section{Evaluation of Global Optimization on 3DPW}
\label{sec:supp:eval_global_opt}

\setlength{\tabcolsep}{5pt}
\begin{table}[ht]
\footnotesize
\centering
% \resizebox{0.9\linewidth}{!}{ 
\begin{tabular}{@{\hskip 1mm}lcc@{\hskip 1mm}}
\toprule
Method & Relative Translation Error & Relative Rotation Error \\ \midrule
Ours w/o Global Optimization &  1.92 & 1.07 \\
Ours (GLAMR) & \textbf{0.66} & \textbf{0.30} \\
\bottomrule
\end{tabular}
% }
\vspace{-2mm}
\caption{Evaluation of our global optimization framework on 3DPW. We evaluate the relative translation error (in meters) and relative rotation error (in angles) between pairs of humans. Here, ``relative'' denotes the relative spatial relationship between two humans.}
\label{table:supp_eval_global_opt}
\end{table}

We also perform experiments on 3DPW with and without our global optimization framework to study the importance of global optimization when there are multiple people in the video. Although 3DPW does not provide accurate GT human trajectories in the global coordinates, the relative translations and rotations between people in 3DPW are quite accurate. Therefore, we compute the relative translations and rotations between pairs of humans and calculate their errors \wrt the ground truth. These metrics, \ie, relative translation and rotation errors, serve as an alternative way to evaluate global reconstruction quality. As shown in Table~\ref{table:supp_eval_global_opt}, using global optimization can greatly reduce the relative translation and rotation errors between humans, which means our global optimization framework can greatly help to reconstruct the spatial relationships of humans in the video.

\section{Effect of Sliding Window Length.}
\begin{figure*}[ht]
    \vspace{-2mm}
    \centering
    \includegraphics[width=0.4\linewidth]{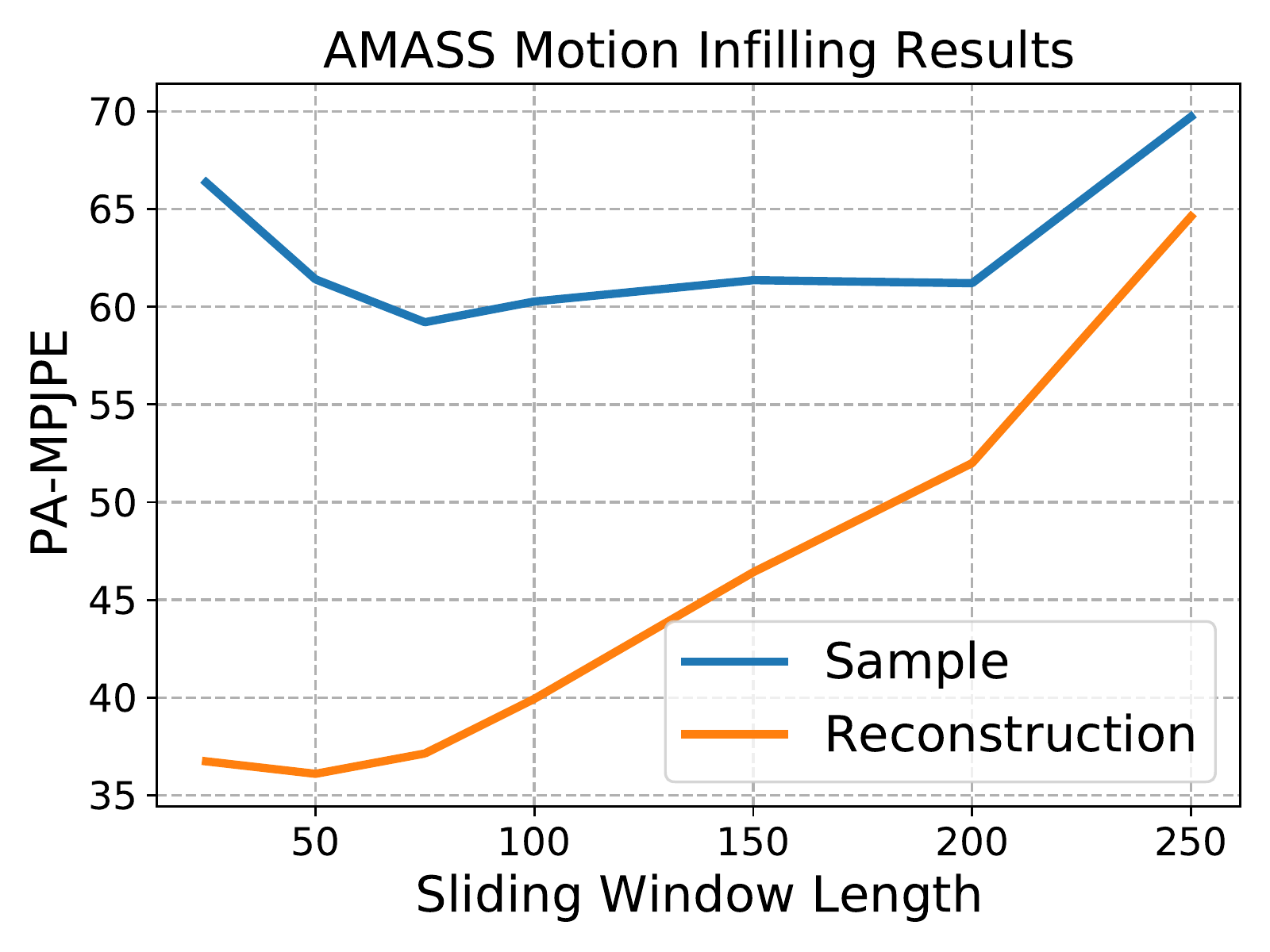}
    \vspace{-2mm}
    \caption{Sample and reconstruction PA-MPJPE \vs sliding window length $h$. The context $h_\texttt{c}$ and look-ahead $h_\texttt{l}$ are always $0.2h$.}
    \label{fig:plot}
    \vspace{-2mm}
\end{figure*}

As shown in Fig.~\ref{fig:plot}, when increasing the window length $h$ (with context $h_\texttt{c}$ and look-ahead $h_\texttt{l}$ being $0.2h$), the reconstruction error increases because it is harder for the latent code $\boldsymbol{z}$ to encode a longer window which contains more motion variations than a shorter window. In the meantime, the sample error first drops and then increases since there is a trade-off: a longer window provides more context for better inference, but it also puts more burden on the latent code as indicated by the increasing reconstruction error.

\vspace{2mm}
\noindent\textbf{Motion Infilling without Visible Pose.}
In the extreme case, when there is no visible pose ($h_\texttt{c}=h_\texttt{l}=0$), our motion infiller can still produce plausible motions sampled from the prior learned from the training motion datasets. In this case, the motion infiller essentially becomes an unconditional VAE model.

\section{Discussion of Limitations}
\label{sec:supp:limitations}
As the first paper on this new problem, our method has a few limitations that are important for future research to address. First, our approach has five stages that are sequentially dependent. Therefore, errors in early stages can propagate to late stages, which may lead to inaccurate global pose estimation. Future work could integrate these stages together to form an end-to-end learnable framework. Second, like many works in human mesh recovery, our approach can only recover the SMPL parameters which omit the fine details of human meshes such as clothing. Integrating neural articulated shapes such as~\cite{deng2020nasa} into our approach could potentially address this problem. Third, our approach is not real-time due to the batch processing and global optimization. Future work could explore a causal version of our approach where only a small window around the incoming frame is optimized, which could substantially improve computational efficiency. Finally, the generative motion infiller and global trajectory predictor in our approach operate for each person independently. Therefore, the generated motions and trajectories may not capture potentially complex and nuanced interactions between occluded people such as hugging or dancing. Future work could address this limitation by employing new generative models that produce interaction-aware motions of multiple people.

\section{Discussion of Potential Negative Impact}
\label{sec:supp:neg_impact}
With its strong ability to reconstruct global human motions and tackle severe occlusions, our method marks a significant step towards global human mesh recovery in the wild. However, misuse of this technology could lead to potential privacy concerns and the propagation of misinformation. For instance, combined with advanced neural rendering approaches~\cite{tewari2021advances}, the reconstructed global human motion of our approach could be used to fabricate videos of human actions that are indistinguishable from real ones. To address this issue, future research should continue to study the detection of synthesized videos with realistic human motion.

\end{document}